\pdfoutput=1

\documentclass[11pt]{article}

\usepackage[]{acl}

\usepackage{times}
\usepackage{latexsym}

\usepackage[T1]{fontenc}

\usepackage[utf8]{inputenc}
\usepackage{todonotes}

\usepackage{microtype}
\usepackage{amsmath,amssymb}
\usepackage{epsfig,graphicx,subfigure,caption}
\usepackage{algpseudocode}
\usepackage[normalem]{ulem}
\usepackage{amsmath}
\usepackage[linesnumbered,algoruled,boxed,noend]{algorithm2e}
\usepackage{xcolor}
\usepackage{color, colortbl}
\usepackage{multirow,booktabs, hhline}
\usepackage{amsmath, bm}
\usepackage[linesnumbered,algoruled,boxed,noend]{algorithm2e}
\usepackage{wrapfig}
\usepackage{listings}
\usepackage{adjustbox}
\usepackage{pifont}
\newcommand{\cmark}{\ding{51}}%
\newcommand{\xmark}{\ding{55}}%

\usepackage{url}
\usepackage[T1]{fontenc}
\usepackage{tikz}
\usepackage{soul}
\usepackage{enumitem}

\setlist[enumerate]{itemsep=0pt, parsep=0pt}
\setlist[itemize]{itemsep=4pt, parsep=0pt}

\usepackage[T1]{fontenc}

\usepackage[utf8]{inputenc}

\usepackage{microtype}

\usepackage{inconsolata}

\usepackage{flushend} 
\usepackage{xspace}
\newcommand{\ours}{\textsc{BotEval}\xspace}

\definecolor{darkgreen}{RGB}{0,100,0}
\definecolor{lightyellow}{RGB}{255,255,224}

\DeclareRobustCommand{\mycircle}[3]{
\tikz[baseline=(char.base)]{
\node[shape=circle, draw, inner sep=0.5pt, minimum size=3mm, fill=#1, text=#2, font=\scriptsize] (char) {#3};}}

\title{\ours: Facilitating Interactive Human Evaluation}

\author{
    Hyundong Cho\textsuperscript{1} \hspace{1em}
    Thamme Gowda\textsuperscript{2} \hspace{1em}
    Yuyang Huang\textsuperscript{1} \AND
    Zixun Lu\textsuperscript{1} \hspace{3.6em}
    Tianli Tong\textsuperscript{1} \hspace{3.2em}
    Jonathan May\textsuperscript{1} \\
\\
\textsuperscript{1}University of Southern California, Information Sciences Institute \\ 
\textsuperscript{2}Microsoft Translator
\\
{\small \texttt{hd.justincho@gmail.com}}
}

\begin{document}
\maketitle
\begin{abstract}

Following the rapid progress in natural language processing (NLP) models, language models are applied to increasingly more complex interactive tasks such as negotiations and conversation moderations. 
Having human evaluators directly interact with these NLP models is essential for adequately evaluating the performance on such interactive tasks. 
We develop \ours, an easily customizable, open-source, evaluation toolkit that focuses on enabling human-bot interactions as part of the evaluation process, as opposed to human evaluators making judgements for a static input. 
\ours balances flexibility for customization and user-friendliness by providing templates for common use cases that span various degrees of complexity and built-in compatibility with popular crowdsourcing platforms.
We showcase the numerous useful features of \ours through a study that evaluates the performance of various chatbots on their effectiveness for conversational moderation and discuss how \ours differs from other annotation tools.

\end{abstract}
\section{Introduction}

As natural language processing (NLP) models become more versatile with the recent advances of language models and their instruction-tuned counterparts~\cite{ouyang2022training}, it is becoming more common to create language agents~\cite{sumers2023cognitive} and apply them to complex interactive tasks, such as negotiations~\cite{chawla2021casino}, conversational moderation~\cite{cho2023language}, reasoning-guided response generation~\cite{zhou-etal-2022-reflect}, and personalized response generation~\cite{liu2023recap}. 

As noted by \citet{smith-etal-2022-human}, the evaluation methodology plays a critical role in accurately comparing models. 
For example, rankings between dialogue models can change depending on whether they are evaluated based on single-turn responses or full conversations.
In addition, \citet{cho2023language} found the evaluators point of view when evaluating a model is also an important factor.  
They showed that human evaluators perceived conversational moderators as more effective in making the evaluators become more cooperative and respectful when the evaluators directly interacted with the moderators while acting as the moderated user (first person point of view) compared to when they evaluated a completed interaction between a moderator and a moderated user as a bystander (third person point of view).  
However, these factors are overlooked in previous approaches that have focused on a simplified evaluation, such as comparing two complete conversations or individual responses~\cite{smith-etal-2022-human, li2019acute}, or specific dialogue applications such as task-oriented dialogue~\cite{cucurnia-etal-2021-matilda, collins-etal-2019-lida}.  
Therefore, it is important to develop evaluation tools that enable an environment that evaluates models in a setting that best encapsulates how humans actually interact with models.  

To facilitate accurate human evaluations of complex interactive tasks, we developed \ours,\footnote{Source code and documentation for \ours can be found at \url{https://github.com/isi-nlp/boteval}. We make the demo video of \ours available at \url{https://justin-cho.com/boteval}. In addition, a live demo of \ours is also available at \url{https://spolin.isi.edu/boteval-dev1} where reviewers can complete a sample human evaluation task.} a comprehensive evaluation toolkit that focuses on enabling human - bot\footnote{We use \textit{bot} loosely to describe any AI system that a human being interacts with.} interactions as part of the human evaluation process. 
For flexibility, it is dynamically configurable to accommodate as many human agents and model agents to interact with each other simultaneously with a custom dialogue manager. 
It is also designed with modular components, such as the interaction interface, instructions, and survey, so that they can be individually adapted to accommodate various use cases.  
While maintaining generalizability, \ours strives to maximize user-friendliness by providing templates for frequent use cases that involve human evaluation where a human evaluator must interact with a NLP model, multiple models, or another human being to measure human performance. 
In addition, it is integrated with Amazon Mechanical Turk (AMT)\footnote{\url{https://www.mturk.com}} for crowdsourcing. 
It can also be deployed independently of these platforms so that it can be used for internal annotations or used with survey tools that allow for external links, such as Qualtrics\footnote{\url{https://www.qualtrics.com}} and Prolific.\footnote{\url{https://www.prolific.com}}

To showcase the usefulness of \ours and demonstrate its key features, we share a case study that uses \ours for evaluating models on their performance on conversational moderation~\cite{cho2023language}. 
In this study, \ours is used to conduct various evaluations: (i) human-bot interactions to compare models; (ii) human-human interactions to measure a human performance threshold; and (iii) completed human-bot interactions by another evaluator to measure evaluator consistency and third-person point of view (POV) results. 

In summary, \ours's main contributions are: 

\begin{itemize}
    \item An open-source and customizable evaluation tool for interactive NLP tasks that incorporates human-bot and human-human interactions into the evaluation process. 
    \item Detailed documentation and templates for various use cases to make modifications easy.
    \item Flexible deployment options with built-in integration with popular crowdsourcing platforms such as AMT and Prolific. 
    \item Evaluation task management features that facilitate task monitoring and managing crowdsource workers. 
    \item Dynamically configurable interaction logic with custom dialogue manager and multi-human and multi-bot evaluation settings. 
\end{itemize}
\begin{figure*}[ht!]
    \centering
    \includegraphics[width=\textwidth]{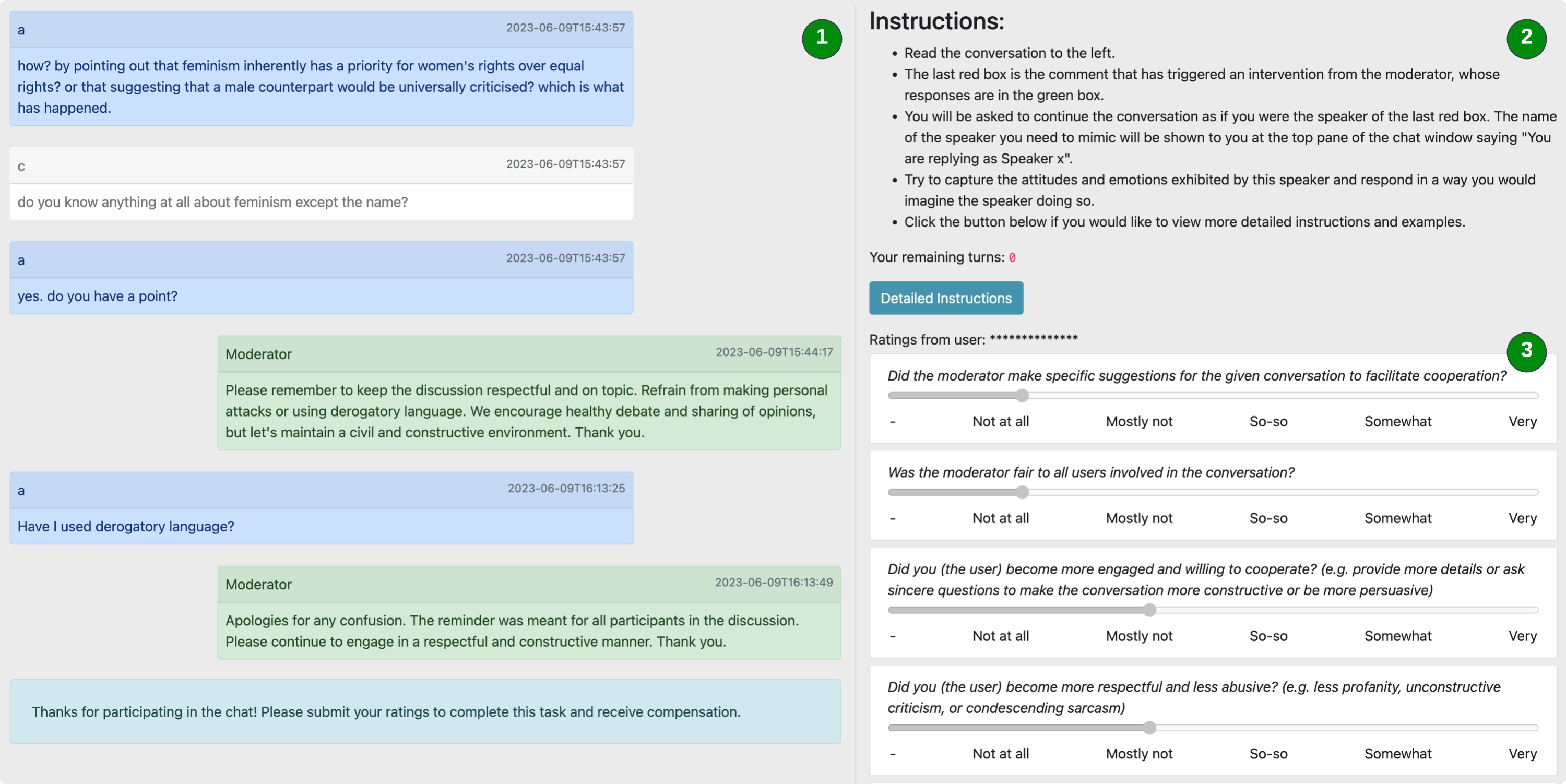}
    \caption{A snapshot of the admin point of view of an evaluation interface with a completed evaluation example. The interface is identical for the evaluator except for the text that shows the evaluator's worker ID (hidden with asterisks in the figure for privacy). 
    The three main components of the user-facing interface are the 
    \mycircle{darkgreen}{white}{1} conversation pane, \mycircle{darkgreen}{white}{2} simple instruction pane, and the  \mycircle{darkgreen}{white}{3} survey pane. 
    }
    \label{fig:boteval_interface}
\end{figure*}

\section{\ours System Overview}

\ours is a web application that provides an evaluation interface, what the human evaluators (i.e., crowdsource workers) see (Section \ref{sec:evaluation_interface}), and an administrator dashboard, what the administrator uses to manage the evaluation task and evaluators (Section \ref{sec:administrator_interface}). 
We recommend that the bots that evaluators interact with are provided as separate APIs that \ours can make queries to, as this isolates the management of the bot deployment and \ours (Section \ref{sec:bot_customization}). 
Human evaluators can be flexibly set to crowdsource workers from AMT or Prolific or any other evaluators with internet access by having them create an account directly for a deployment of \ours using a public external link. 
An evaluation task is configured with a central YAML config file that identifies the frontend components to use, the deployment environment, and the crowdsourcing platform to use. 


\subsection{Evaluation interface}
\label{sec:evaluation_interface}

A sample evaluation interface for the case study later described in Section \ref{sec:case_study} is shown in \autoref{fig:boteval_interface}. 

The evaluation interface consists of three main components: \mycircle{darkgreen}{white}{1} \textit{Conversation pane}: a section where the interaction between the human and the bot takes place. This pane can be easily customized to contain seed conversations to serve as initial starting points for interactions to continue off of or it can instead contain any piece of text or completed conversation without requiring any interactions from the evaluators, making \ours also suitable for simpler annotation tasks.  
\mycircle{darkgreen}{white}{2} \textit{Instruction pane}: this is an optional section that shows the main directions. Evaluators can see detailed instructions by clicking on the detailed instructions button. Administrators can choose to show detailed instructions as part of the consent form if one is needed to make sure that evaluators have read them. 
\mycircle{darkgreen}{white}{3} \textit{Survey pane}: this is where the human evaluators provide their evaluations. In the given example, it is configured to only be shown after the human evaluators have interacted with the bot for a set number of turns. 

The conversation pane and instruction pane is configurable by providing custom HTML scripts, while the survey pane is even more easily customizable by configuring a YAML config file. An example of the YAML config file is shown in Appendix \ref{sec:sample_survey_config}.
An optional consent form can be shown to evaluators as well, which is also managed with a separate HTML file. 
Further detail on how the consent form can be configured is in Appendix \ref{sec:consent_form_config}.

\subsection{Administrator dashboard}
\label{sec:administrator_interface}

\begin{figure*}[ht!]
    \centering
    \includegraphics[width=\textwidth]{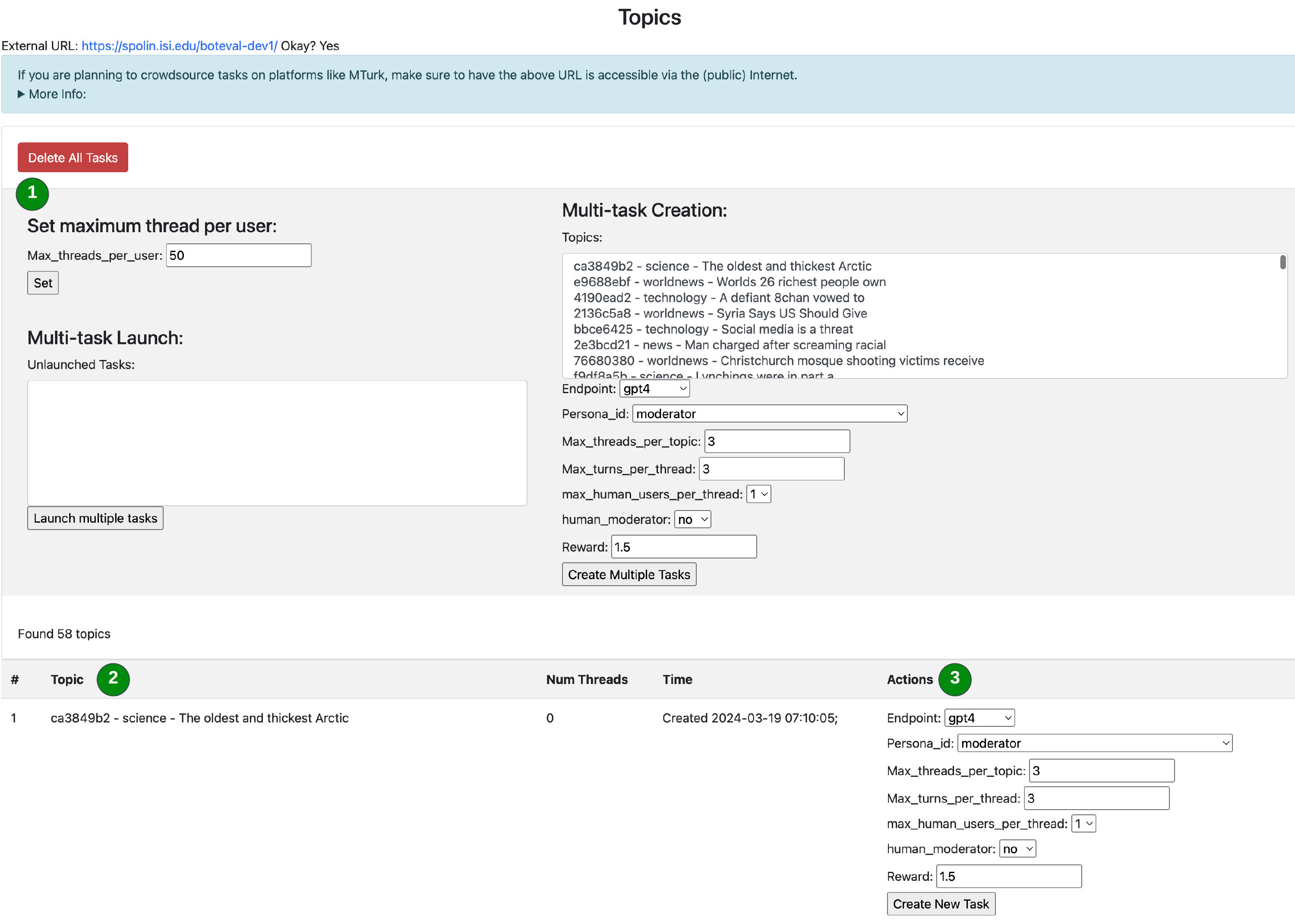}
    \caption{A snapshot of the topics page of the admin dashboard. 
    \mycircle{darkgreen}{white}{1} is a parallel management tool that enables setting global configurations such as how many tasks each evaluator is allowed to complete and launching or deleting multiple tasks at once. \mycircle{darkgreen}{white}{2} is a topics table that shares more information about each topic, such as its name, how many tasks have been created, and when they were created. \mycircle{darkgreen}{white}{3} is a list of parameters that can be chosen for launching a task, which includes parameters that can be passed on to API queries for the bots.  
    }
    \label{fig:boteval_admin_dashboard}
\end{figure*}

\ours's administrator dashboard provides numerous features for managing evaluation tasks and evaluators. 
Its main benefit is a GUI that enables a non-technical user to easily become an administrator for human evaluation tasks.
The topics page, shown in \autoref{fig:boteval_admin_dashboard}, is one view that allows the management of launching and deleting tasks. 
A topic refers to any predetermined context, such as seed conversation or external information relevant for the evaluation task. 
These topics are provided to \ours as a JSON file.
If a user is interested in general open-domain dialogue evaluation, measuring a model's general conversational capabilities, they can use a dummy topic file that contains an empty dictionary. 
This will launch an evaluation task that starts a conversation from scratch, with the human evaluator initiating the first turn. 

After launching tasks, users can use the administrator dashboard to conveniently examine tasks that are completed or in progress with the same interface that the evaluators used to complete the task to easily visualize their work rather than examining a database or JSON file, as shown in \autoref{fig:boteval_interface}. 
The user can also directly export individual JSON files of the collected data if needed. 
Also, tasks can be deleted in batches using the parallel management tool shown in  \mycircle{darkgreen}{white}{1} of \autoref{fig:boteval_admin_dashboard}.

In addition to these features, we provide convenient AMT-specific features 
for managing workers and tasks known as human intelligence tasks (HITs). 
One of the most convenient features is being able to directly assign and remove qualifications for workers after examining their work without having to leave the administrator dashboard. 
This is an important convenience feature for ensuring the quality of work for human evaluations are kept to the desired standard by blocking unreliable workers. 
Another is being able to make bonus payments directly after examining the completed task, which is useful when each task is expected to involve variable rewards, such as to account for each HIT taking a different amount of time to complete.

\subsection{Bot customization}
\label{sec:bot_customization}

Users are given multiple options to choose how they will service the bot that they want to evaluate, but the recommended setup is to set up a separate RESTful API and defining a logic within \ours to interface with this API. 
As shown in \mycircle{darkgreen}{white}{3} in \autoref{fig:boteval_admin_dashboard}, users can define task-specific parameters for bots that get passed on to the API if the API allows for it. 
This is useful if you are using the same model but adjusting the instruction prompt (e.g., using OpenAI endpoints).
While \ours users have the option to launch bots simultaneously on the same server with \ours's process, it is more efficient to separately manage human evaluation tasks and the processes that load and query NLP models because most NLP models are better served with GPUs for reducing latency. 



\subsection{Sourcing human evaluators}

\ours can be customized to use with any crowdsourcing platform, and it is designed to be directly used with many popular ones such as AMT, Prolific, and Qualtrics. 
If the goal is to do internal annotations, the setup is even simpler as the user only has to configure \ours to not use any. 
Then the user can share their custom URL with the evaluators, where they can sign up and directly work on tasks that are made available to them without going through any other platform. 

\section{System Architecture}

\begin{figure}
    \centering
    \includegraphics[width=\columnwidth]{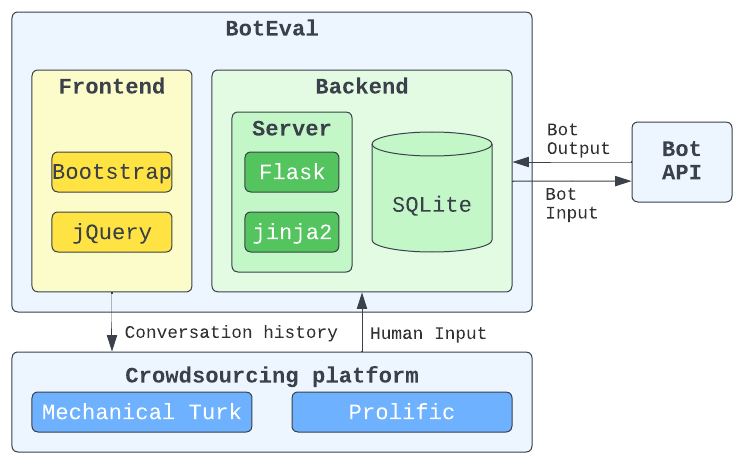}
    \caption{\ours system architecture. We use popular frameworks that are well documented and easy to use.}
    \label{fig:boteval_architecture}
\end{figure}

An overview of \ours's system architecture is shown in \autoref{fig:boteval_architecture}.  
\ours is a web application (i.e., a client-server model). 
We describe the front- and back-end technology stacks in the following sections. 

\subsection{Frontend}

The frontend is a simple web interface (i.e., HTML) created with Bootstrap stylesheet.
While the majority of the HTML structure is constructed on the server side using Jinja2, some dynamic updates such as responses coming from bots or other participants in the interaction are achieved using AJAX and RESTful APIs.

\subsection{Backend}
The backend is implemented in Python language using Flask framework, following a model-view-controller architecture pattern.
\textit{Models} are implemented using Python classes and stored in a relational database, specifically SQLite. 
In addition, we use SQLAlchemy, an object-relational mapper, to abstract the mapping between Python classes and database tables.
For \textit{views}, Flask uses Jinja2 for server side templating of HTML pages.
\textit{Controllers} are based on Flask's builtin URL routers and RESTful API constructs.

While internally our server is an HTTP server, crowdsourcing platforms such as AMT require annotation interface be served via secure connections (HTTPS).
HTTPS can be enabled by obtaining and installing an SSL/TLS certificate.
We use free certificates from Certbot,\footnote{\url{https://certbot.eff.org/}} and configure Nginx\footnote{\url{https://nginx.org/en/}} as a reverse proxy server for HTTPS requests.

Some scenarios may require several simultaneous instances of \ours to facilitate multiple annotation tasks, and obtaining SSL certificate for each instance maybe cumbersome. 
We address this problem by using a different TCP port for each instance, and configuring a single Nginx (with SSL certificate) route requests for all instances.

\section{Case Study and Use Cases}

\subsection{Case study: conversational moderation evaluation}
\label{sec:case_study}

To showcase the usefulness of \ours, we share a case study that uses \ours to conduct a study on how effective various zero-shot instruction-tuned language models (ITLM) and dialogue model are in performing \textit{conversational moderation} (CM) ~\cite{cho2023language}.\footnote{The \ours template for this work is available at \url{https://github.com/isi-nlp/isi_darma/tree/main/boteval-darma-task}.} 
Instead of iron-fisted approaches to moderation such as deleting comments or banning users, which may exacerbate societal polarization as these users find refuge in echo chambers, CM seeks to have moderators interact with users exhibiting problematic behavior to guide them back to more constructive and respectful conversations. 

This study makes full use of \ours~as it requires evaluating multiple bots by interacting with them for a preset number of turns (in this case 3), starting with a variety of conversation stubs. 
The evaluations were conducted with all desired configurations simultaneously to get the most representative and fair results that is not affected by any confounding factors such as recency bias. 
The evalution was conducted with AMT, and being able to easily monitor evaluations enabled rapid iterations of updating the instructions and giving feedback to the evaluators. 

Therefore, \ours was integral in being able to refine the evaluation study efficiently and ultimately collect statistically meaningful results for an interactive evaluation setup.
The study showed that prompt-engineered ITLMs outperformed prosocial dialogue models and that a conflict resolution prompt based on the Socratic method was the best performing prompt. 
In addition, one of this work's central findings was discovering that there are differences between evaluation results when the models were evaluated from a first person point of view (POV) and a third person POV. 
With \ours, collecting human evaluations in these two different settings was a simple change of updating the topics file such that the conversation stubs were the completed conversations, rewording the questions such that it is in third person POV, and setting the number of turns required for human evaluators to interact with the bots to zero. 

\begin{table*}[t!]
\begin{adjustbox}{max width=\linewidth}
    \centering
    \begin{tabular}{lcccl}
        \toprule 
        Name &  Human-bot interaction-focused &  Crowdsource integration &  Multi-human \& bot support & Language   \\ \midrule

        \textbf{\ours} (Ours) & \cmark & \cmark & \cmark & Python \\ 
        Mephisto \cite{mephisto} & \xmark & \cmark & \xmark & Python \\ 
        Pigeon$^{11}$ & \xmark & \xmark & \xmark  & Python  \\ 
        MATILDA \cite{cucurnia-etal-2021-matilda} & \xmark & \cmark & \xmark & Python \\ 
        LIDA \cite{collins-etal-2019-lida} & \xmark & \cmark & \xmark & Python \\ 
        INCEpTION \cite{klie2018inception} & \xmark & \xmark & \xmark & Java \\ 
        GATE \cite{cunningham2002gate} & \xmark & \xmark & \xmark & Java \\ 
        BRAT \cite{stenetorp2012brat} & \xmark & \xmark & \xmark & Python \\ 
        doccano~\cite{doccano}  & \xmark & \xmark & \xmark & Python \\ 
        Potato~\cite{pei2022potato} & \xmark & \xmark & \xmark & Python \\ 
        Argilla$^{10}$ & \xmark & \xmark & \xmark  & Python  \\ 
        Prodigy$^{12}$ & \xmark & \xmark & \xmark  & Python  \\ 
        DialogueView \cite{yang2005dialogueview} & \xmark & \xmark & \xmark & TcK/TK \\ 
        DART \cite{weisser2016dart} & \xmark & \xmark & \xmark & Perl \\
        Anvil \cite{kipp2001anvil} & \xmark & \xmark & \xmark & Java \\ 
        EZCAT \cite{guibon-2022-ezcat} & \xmark & \xmark & \xmark & Javascript \\ 
         \bottomrule
    \end{tabular}
\end{adjustbox}
    \caption{Comparison overview with other annotation tools. \ours~innately supports evaluations that require human-bot interactions and allow for multiple human agents or bot agents to be involved in each evaluation sample.}
    \label{tab:comparison}
\end{table*}

\subsection{Main use cases}

\ours's main differentiation with previous annotation tools and frameworks is that it is focused on, but not limited to, interactive use cases. 
In other words, it is useful when the annotated data is not static, e.g., bot responses over multiple turns or other dynamic outputs that can change based on user interaction.
Therefore, \ours is  appealing for evaluating or collecting data for conversational tasks that usually require multi-turn interactions for fulfilling the goal, rather than a single generated output. 
Many real-life tasks go through multi-turn interactions, such as negotiations~\cite{chawla-etal-2021-casino}, counseling~\cite{mehta-etal-2022-psychotherapy}, and improvisational theater~\cite{cho-may-2020-grounding}.  
As artificial systems become more capable, more will be applied to completing these complex multi-turn tasks, and \ours will serve as a handy starting point for facilitating their evaluation. 

Although \ours was designed for interactive tasks, \ours can also be easily adapt for simple static annotation tasks by simplifying the conversation pane in \autoref{fig:boteval_interface}. 
This pane can contain any other modality such as images, video, and audio, and adjusting the instruction and survey panes accordingly can make \ours also suitable for text classification or conversation-level or turn-level comparisons, similar to ~\citet{smith-etal-2022-human}. 
As \ours gets actively used for more research studies, we will be able to provide a variety of templates that accommodate a comprehensive set of use cases, further lowering the effort required to conduct effective human evaluation for new studies.

\section{Related Work}
\label{sec:related_work}

In \autoref{tab:comparison}, we compare \ours~with other related annotation tools and discuss differences further here. 

\subsection{General text annotation tools}
\label{sec:related_work/general_annotation_tools}

A popular general annotation tool is Mephisto \cite{mephisto}, which started by isolating the crowdsourcing features from ParlAI~\cite{miller2017parlai}. 
Mephisto provides a general annotation framework that interfaces with Amazon Mechanical Turk and Prolific and includes basic templates for simple annotation tasks. 
\ours adapted many of its AMT integration features, but Mephisto is not customized for common interactive data annotation and evaluation use cases, and thus requires nontrivial effort to create a human evaluation environment for interactive NLP tasks where a human evaluator needs to interact with a bot or another human and then evaluate their performance. 
ParlAI still provides templates for Mephisto for human-bot interactions\footnote{\url{https://parl.ai/docs/tutorial_crowdsourcing.html}}, but it is not easy to use with a dialogue model that is not developed with ParlAI. 
With \ours, we also provide a GUI administrator dashboard for task and worker management, which is absent in ParlAI and Mephisto. 

GATE~\cite{cunningham2002gate} and INCEpTION~\cite{klie2018inception} are annotation tools that provide many predefined features, but they are also not designed for interactive human evaluations. 
Other simpler general text annotation tools that share similar limitations are Doccano~\cite{doccano}, brat~\cite{stenetorp2012brat}, Argilla\footnote{\url{https://argilla.io}}, Potato~\cite{pei2022potato} and 
Pigeon\footnote{\url{https://github.com/agermanidis/pigeon}}, which are web-based annotation tools that enable rapid annotations for text classification and machine translation.
Prodigy\footnote{\url{https://prodi.gy}} is a commercial annotation tool for text annotations that provides similar features.

\subsection{Dialogue annotation tools and evaluation methodologies}
\label{sec:related_work/dialogue_annotation_tools}

A prominent set of annotation tools specific to dialogue are centered around task-oriented dialogue~\cite{budzianowski-etal-2018-multiwoz}. 
LIDA~\cite{collins-etal-2019-lida} is an annotation tool that provides useful features for efficiently making turn-level annotations, incorporating model-provided label recommendations to speed up annotations, and resolving inter-annotation disagreements.
MATILDA~\cite{cucurnia-etal-2021-matilda} builds on LIDA for multilingual support and improved management of crowdsourcing tasks among multiple workers. 
However, they do not have built-in compatibility with popular crowdsourcing platforms and do not support human-bot interactions to take place within the crowdsourcing task. 
A lightweight option for dialogue annotations is
EZCAT~\cite{guibon-2022-ezcat}, which provides a web-based serverless annotation framework that focuses on enhanced accessibility for conversation-level and turn-level annotations. 

Other work have created tools for multimodal annotations or speech-based annotations. 
Anvil~\cite{kipp2001anvil} provides a multi-modal dialogue annotation tool that enables annotation of audiovisual content. 
DialogueView~\cite{yang2005dialogueview} is an annotation tool that is focused on segmenting audio conversations. 
DART~\cite{weisser2016dart} focuses on enabling efficient annotations of speech acts and linguistic criteria to facilitate corpus-based research into pragmatics. 
Text is still the primary focus of \ours~and the templates we provide, but \ours remains general enough to be adapted to such cases as well by modifying the templates we provide. 

\section{Conclusion}

We presented \ours and its usefulness in collecting human evaluations for interactive tasks that require live human-bot interactions through a case study of evaluating various language models on their ability to conversationally moderate online discussions. 
\ours provides a customizable interface that can be adapted for various evaluation and annotation use cases while also providing integration with popular crowdsourcing platforms and task management features. 
We hope that this work will serve as an important foundation for setting up custom interactive human evaluation tasks that facilitate our understanding of more complex NLP systems as they become increasingly sophisticated and capable. 

\section*{Limitations}

We designed \ours to be modular such that 
 customizing existing templates and modifying the dialogue manager's logic is simple, but it is yet not configured so that the task management process, shown in \autoref{fig:boteval_admin_dashboard} is independent of the process that serves the evaluation, shown in \autoref{fig:boteval_interface}. 
This means that any updates to \ours that help with task management cannot be applied without restarting evaluation tasks that were launched already, which will interfere with any concurrent tasks that evaluators are working on. 
While inconvenient, this has not been a major issue as restarting can be done quickly such that it does not interfere the work of many evaluators and this doesn't mean that existing crowdsourcing tasks, as in the case of AMT, will be deleted and need to be relaunched again. 

Another challenge for using \ours may arise from the difficulty of managing a separate process that serves the bots that the human evaluators will interact with. 
However, if the \ours user is able to launch a bot as part of \ours, refactoring the code for that bot such that its responses are accessed through an API instead is a simple modification with plenty of online tutorials and tools, such as FastAPI.\footnote{\url{https://github.com/tiangolo/fastapi}}

\section*{Acknowledgements}
This material is based upon work supported by the Defense Advanced Research Projects Agency
(DARPA) under Agreement Nos. HR00112290025 and HR00112490374. Approved for public release; distribution is unlimited.

\bibliography{anthology,custom}

\appendix

\section*{Appendix}
\label{sec:appendix}

\section{Evaluation interface configurations}

\subsection{Sample survey configuration}
\label{sec:sample_survey_config}

\begin{figure*}[ht!]
    \centering
    \includegraphics[width=\textwidth]{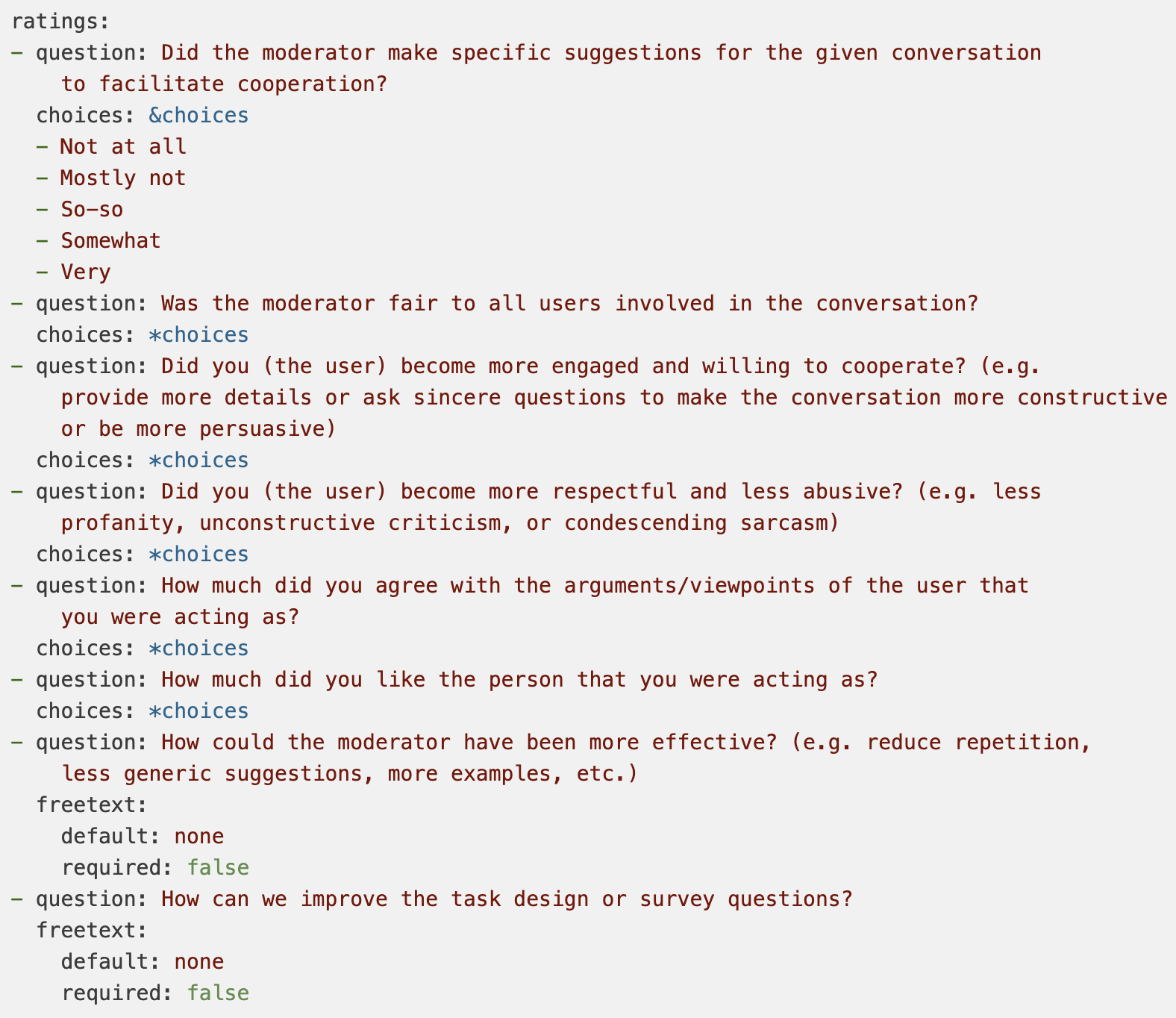}
    \caption{An example of the survey pane configuration that contains a custom Likert scale and freeform text input fields. This configuration corresponds to the survey pane partially shown in \autoref{fig:boteval_interface}. 
    }
    \label{fig:survey_pane_config_example}
\end{figure*}

An example survey configuration is shown in \autoref{fig:survey_pane_config_example}. 
Survey components are also easily dynamically configurable with common input options such as radio buttons, Likert scales, freeform text, etc. 
Users can create their own survey with HTML as well, but the customization options we provide through the YAML file covers most conversation-level evaluation use cases. 

\subsection{Consent form configuration}
\label{sec:consent_form_config}

\begin{figure}[ht!]
    \centering
    \includegraphics[width=\columnwidth]{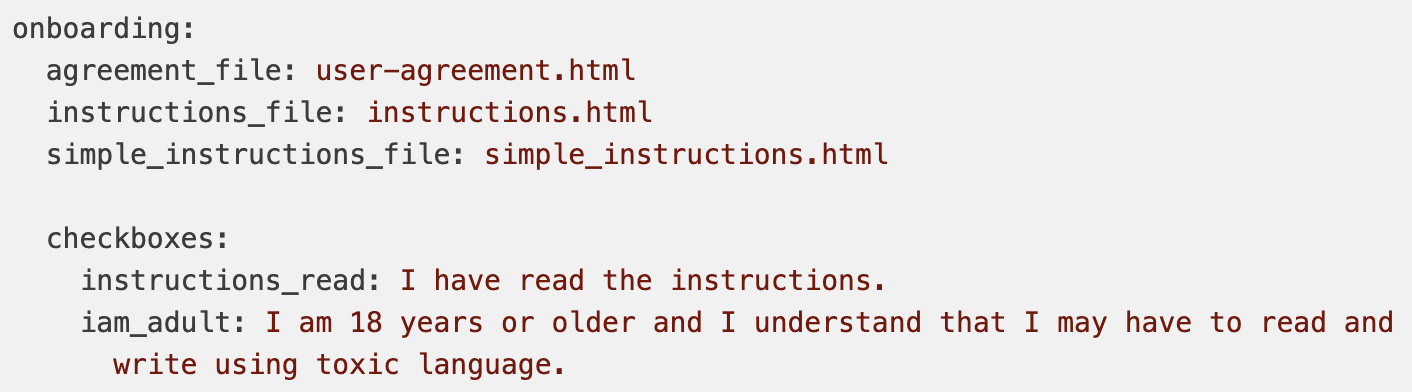}
    \caption{An example of configuring the consent form. The \texttt{agreement\_file} parameter should point to the HTML file that shows the content of the consent form. 
    }
    \label{fig:consent_form_configuration}
\end{figure}

An example of configuring the consent form is shown in \autoref{fig:consent_form_configuration}. 
When deployed without a crowdsourcing platform, these appear as checkboxes in the sign up process. 
Within AMT, we automatically sign up the workers with their worker ID and do not require a password, but they have to check the same checkboxes in order to move on to the task if they are doing the task for the first time.

\end{document}